\ifx\XeTeXversion\undefined\pdfoutput=1\fi
\documentclass[10pt,twocolumn,letterpaper]{article}

\usepackage[pagenumbers]{cvpr} % arXiv version: authors shown, page numbers

\usepackage{multirow}
\renewcommand{\paragraph}[1]{\par\noindent\textbf{#1}}
\usepackage{listings}
\lstdefinestyle{prompt}{
  basicstyle=\ttfamily\scriptsize,
  breaklines=true,
  breakindent=0pt,
  columns=fullflexible,
  keepspaces=true,
  frame=single,
  framesep=3pt,
  rulecolor=\color{gray!60},
  xleftmargin=0pt,
  aboveskip=3pt,
  belowskip=8pt,
}

\definecolor{cvprblue}{rgb}{0.21,0.49,0.74}
\usepackage[pagebackref,breaklinks,colorlinks,allcolors=cvprblue]{hyperref}

\def\paperID{*****} % *** Enter the Paper ID here
\def\confName{CVPR}
\def\confYear{2027}

\title{CoaG: Cylinders on a Grid for Coarse 3D Layout Control in Video Generation}

\author{Zhangsihao Yang$^{*}$\\
{\tt\small zshyang1106@gmail.com}
\and
Mengyi Shan$^{*}$\\
University of Washington\\
{\tt\small shanmy@cs.washington.edu}
}

\begin{document}
% Title block and the teaser figure span both columns; \maketitle's own \twocolumn is disabled inside.
\twocolumn[{%
  \renewcommand\twocolumn[1][]{#1}%
  \maketitle
  % Teaser: Fig. 1 sits under the title, full width (see main.tex).
\begin{center}
  \vspace{-1.2em}
  \includegraphics[width=\textwidth]{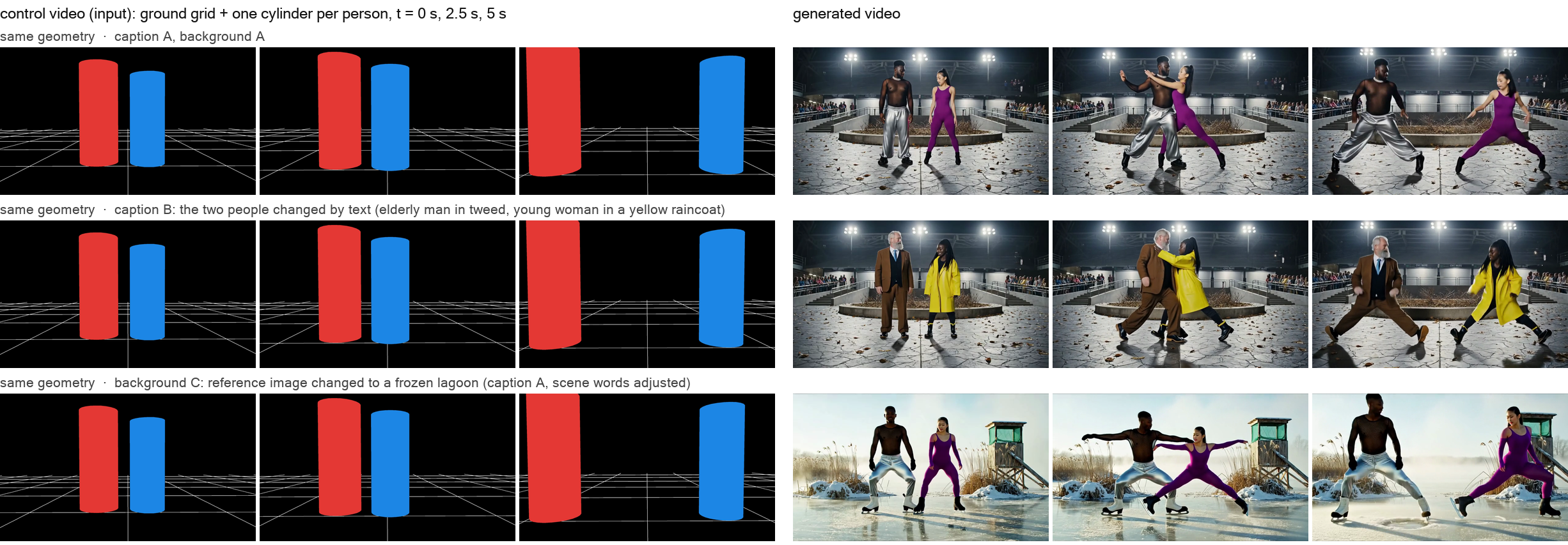}
  \captionof{figure}{\textbf{A ground grid and one cylinder per person are enough to direct a video model.} Left: the control video (three of its 81 frames): a ground grid and two cylinders drawn on it, seen from a static camera. Right: videos generated from that same control video under three settings. Row 1, caption A with background A: the text describes two dancers, a man in silver trousers and a woman in a magenta bodysuit, and the reference image is the stadium plaza. Row 2, caption B with background A: the text now describes an elderly man in a tweed suit and a young woman in a yellow raincoat; everything else is unchanged. Row 3, caption A with background C: the reference image is replaced by a frozen lagoon (the scene words of the caption are adjusted to match, the people and the action are those of caption A). In all rows the people stand where the cylinders stand and keep their left-to-right order and height; the text decides who they are and what they do, the reference image decides where they are.}
  \label{fig:teaser}
  \vspace{0.8em}
\end{center}

}]
{\renewcommand{\thefootnote}{}\footnotetext{$^{*}$Equal contribution.}}
\begin{abstract}
We ask how little geometry a person has to draw to control both where people stand and where the camera moves in a generated video. Our answer is a ground plane and one cylinder per person. A user draws a grid on the ground, places one cylinder where each person should stand, moves the cylinders and the camera over 81 frames, and the model renders a photoreal video in which the people occupy the cylinders' positions, move as the cylinders move, and are seen from the drawn camera. Appearance comes from a text prompt and a background reference image; layout and motion come from the geometry. Because no dataset pairs such a signal with video, we build the pairs ourselves: an automatic engine writes 2000 captions from a combinatorial seed, generates a clip for each with a text-to-video model, and lifts every clip back to its geometry with person tracking, background inpainting, an agentic ground-mask loop, feed-forward multi-view reconstruction and a plane fit, with no real footage and no manual labels. A LoRA on Wan2.2-Fun-Control trained on 1935 such tuples follows drawn layouts and camera paths on hold-out clips: the generated people match the cylinders' count, order, position and height, the text changes who they are, the reference image changes where they are, and dolly-in, orbit, pan and crane paths are followed, dolly-out only weakly.
\end{abstract}

\section{Introduction}
\label{sec:intro}

\begin{figure*}[t]
  \centering
  \includegraphics[width=\textwidth]{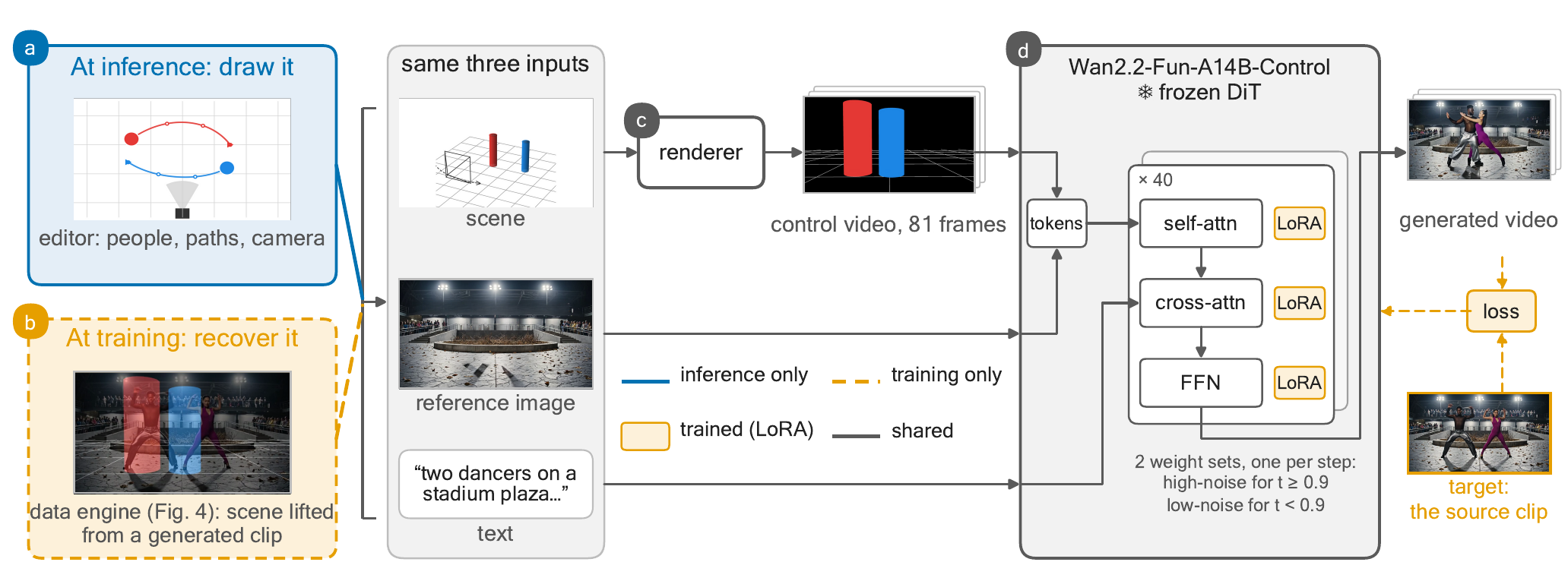}
  \caption{\textbf{Overview.} Top, inference: a person authors the layout in the editor (cylinders on the grid, their paths, a camera path); the rendered control video, a reference image for the background and a text prompt go into Wan2.2-Fun-A14B-Control, which is frozen except for one LoRA per expert; the output video places the people where the cylinders are and moves the camera as drawn. Bottom, training: a language model writes captions from seed cards, a text-to-video model turns them into clips, and the data engine (Fig.~\ref{fig:engine}) lifts each clip to people, background, ground plane and cameras and renders the same control video; the tuple (control video, background, caption, target clip) trains the LoRAs. The renderer and the three input slots are identical at training and at inference; only the control video's origin differs.}
  \label{fig:overview}
\end{figure*}

Control signals for video generation are either dense or blind to people. Pose skeletons, depth maps, edges and tracked point videos steer a model precisely, but the user must already own a video that produces them, which is the video they wanted to make. Camera-control methods move the viewpoint along an authored path, but say nothing about where the people in the scene stand or how they move. Layout methods place boxes in the image plane, so a box drawn for a person in the distance and a box for a person nearby are just two rectangles; the ground that ties the two together is lost, and so is the camera.

We take the opposite starting point and ask for the least geometry a person can author by hand that still fixes both the subject layout and the camera. Our answer is a ground plane, drawn as a grid, and one upright cylinder per person standing on it (Figure~\ref{fig:teaser}). A cylinder has a footprint on the plane, a height and a color; moving it over time gives a trajectory; moving the camera gives a shot. Everything else, the identity and clothing of the people, the place, the light, comes from a text prompt and from a background reference image, which is where a user wants freedom. The representation is coarse on purpose: it can be drawn in a minute, it carries no pose that the user would have to animate, and it is unambiguous about the two things dense signals and 2D layouts get wrong, the ground and the depth order.

Two developments make this practical. Open video models~\cite{wan2025,cogvideox2024} now ship with control adapters (Wan2.2-Fun-Control~\cite{videoxfun2025}) that accept an arbitrary control video and a reference image, so a new control modality can be taught with a LoRA on a few thousand clips instead of a new model. And text-to-video models are good enough to be the data source: instead of collecting footage in which people stand on known ground under known cameras, we generate the footage from captions we control and recover the geometry from it.

Our contributions are:
\begin{itemize}
  \item The plane-and-cylinders control representation and its rendering specification (Sec.~\ref{sec:control}).
  \item An automatic data engine that turns text-to-video output into aligned (control video, background image, caption, target video) tuples, with no real footage and no manual labels (Sec.~\ref{sec:engine}).
  \item A LoRA that follows drawn layouts and camera paths, with an authoring demo (Sec.~\ref{sec:training}, \ref{sec:experiments}).
\end{itemize}

\section{Method}
\label{sec:method}

The method has three parts (Figure~\ref{fig:overview}). The control representation (Sec.~\ref{sec:control}) defines what the user draws and how it is rendered into a control video. The data engine (Sec.~\ref{sec:engine}) produces training tuples in which the control video was not drawn but recovered from a generated clip. Training and authoring (Sec.~\ref{sec:training}, \ref{sec:authoring}) fine-tune an open control model on those tuples and give the user an editor whose output is rendered by the same code.

\begin{figure*}[t]
  \centering
  \includegraphics[width=\textwidth]{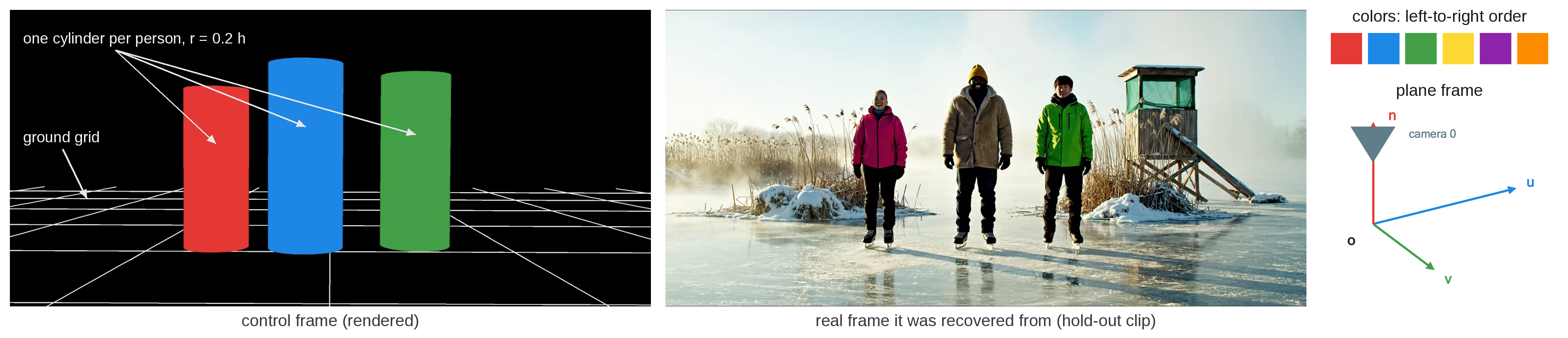}
  \caption{\textbf{The control video.} One rendered frame with its elements called out, and the real frame it was recovered from. The ground is a wireframe grid drawn in the plane frame, whose origin is the foot of the first camera on the plane, $u$ the first camera's forward direction projected onto the plane, $n$ the plane normal and $v = n \times u$; the grid spacing is the median subject height, so all lengths are in subject heights. Each main person is one solid cylinder of radius $0.2h$ and a height $h$ fixed per clip from the first second, standing on the lifted foot point. Colors follow the left-to-right order of the people in the first frame from a fixed six-color palette; cylinders are drawn far to near on a black background with no lighting; the video has 81 frames at 16 fps under the clip's camera (the authored one at inference).}
  \label{fig:controlspec}
\end{figure*}

\subsection{Control representation}
\label{sec:control}

The control signal is a video of 81 frames at 16 fps, the base model's length and rate, rendered on a black background from a scene with exactly two kinds of objects.

\paragraph{Ground grid.} The ground is one plane, drawn as a wireframe grid of white lines. The grid lives in the plane's own coordinate frame: the origin is the point on the plane directly under the camera of the first frame, the first in-plane axis is the camera's forward direction projected onto the plane, and the second axis is the cross product with the plane normal. The grid spacing equals the median height of the people in the clip, so one grid cell is one person tall. The grid is clipped to the padded convex hull of the visible ground, so it ends roughly where the real ground ends. We chose a grid rather than a filled plane because a grid under perspective gives the model two things a filled region does not: the scale of the scene, from the cell size, and the camera motion, from how the lines slide between frames.

\paragraph{Cylinders.} Each main person is one solid upright cylinder standing on the plane. Its radius is 0.2 times its height. Its height is fixed for the whole clip and set from the first second, when every main person is fully visible with their feet on the ground. Its color comes from a fixed palette of six colors, assigned by left-to-right order in the first frame, so the same person keeps the same color for the whole clip and colors never repeat within a clip. Cylinders are drawn back to front, so a nearer cylinder covers a farther one. A solid colored cylinder keeps identity and occlusion order unambiguous when people cross, which a wireframe would not; a constant height turns the cylinder into a pure position-and-motion signal, because the only thing that changes about it is where it stands. The rendering constants are listed in Figure~\ref{fig:controlspec} next to a rendered frame and its real frame.

\subsection{Data engine}
\label{sec:engine}

The control signal is new, so no dataset pairs it with video. We build the pairs ourselves from a text-to-video model. The design question is coverage: the training set must be diverse exactly where a user is free at inference time, the place, the appearance of the people, the action and the camera, while every clip must still be recoverable as one ground plane with people standing on it. Both requirements are enforced upstream, in the captions, rather than filtered downstream, in the reconstruction.

\begin{figure*}[t]
  \centering
  \includegraphics[width=\textwidth]{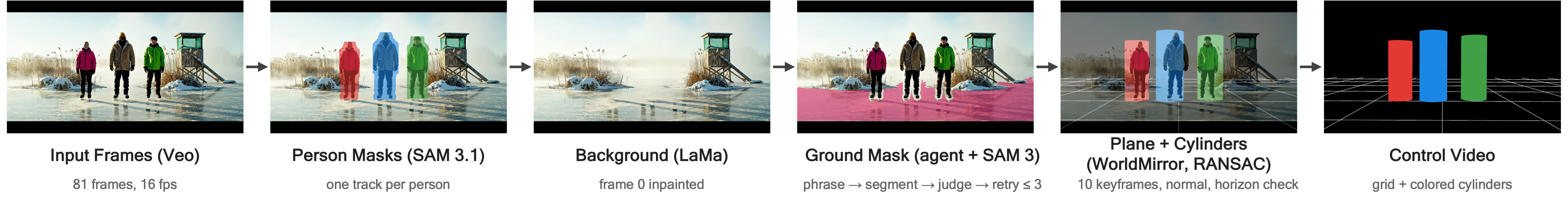}
  \caption{\textbf{The data engine.} Every training tuple is produced automatically from text. A language model writes the caption from a combinatorial seed card (region, ground archetype, action family, camera move, weather or lighting, number of people, crowd flags), and a second model validates and repairs it; a text-to-video model turns the caption into the clip. The annotation system then lifts the clip to geometry: SAM 3.1 tracks the people, LaMa removes them from frame 0 to give the background reference image, an agent loop proposes and judges SAM 3 phrases for the ground, HunyuanWorld-Mirror recovers cameras, points and normals on 10 keyframes (shown as a 3D view with the camera frusta), a RANSAC plane is fitted and checked against the GeoCalib horizon, each person's feet are lifted onto the plane, and the grid and cylinders are rendered into the control video. The tuple (control video, background image, caption, target clip) is what the LoRA is trained on; no real footage and no manual labels enter the pipeline.}
  \label{fig:engine}
\end{figure*}

\paragraph{Captions.} Asking a language model to write two thousand diverse captions collapses into clich\'es after a few hundred, and coverage cannot be proven afterwards. A pure template, a fixed list of scenes crossed with a fixed list of actions, proves coverage but caps the number of places at the length of the list. We use seeds instead: every caption receives a card dealt from a combinatorial grid, so coverage is guaranteed by construction, and the language model only invents the specifics inside the card. The writer must name a concrete place and a concrete action that fit the card; it never copies the card's words. The card has the following axes. A region (50 hand-written entries across continents, climates, city and countryside) sets architecture, vegetation, light and clothing. A ground archetype (162 entries, from a granite plaza and a container-port quay to a boxing ring, a frozen lake and a coffee-drying patio) sets the flat surface underfoot; the dealer never repeats a (region, archetype) pair, so every caption describes a different kind of place in a different part of the world. An action family (118 entries from ball sports, combat, dance, athletics, acrobatics, street games, fitness drills and surface-specific work) is a direction, not a script; the writer produces a sibling move or a variation in the same spirit. The camera is one of eight predefined moves: static tripod, pan, orbit, dolly-in, dolly-out, handheld circle, crane down, sideways track. Outdoor cards draw one of 12 weather and time-of-day states, indoor cards one of 6 lighting states. The environment is either still (40 percent of the cards) or carries one of 14 motion elements such as wind, rain, steam, confetti or light trails. A motion range (in place, short range, long range) is dealt in a 30/40/30 ratio, a person count from six buckets (500, 450, 350, 300, 200 and 200 cards for one to six people), and two flags: a crowd off the playing surface on 20 percent of the cards where an audience is natural, and a subject approaching the camera after the first second on 15 percent. Because the axes are dealt from shuffled decks with exactly the intended counts rather than drawn independently, every share is exact, and the whole dataset is reproducible from the vocabularies and one integer seed.

Not every card can be made to work. A basketball needs a hard floor, a sprint needs room, a shot put needs open sky, ice takes only skating-type actions, a small venue holds at most three people. These eligibility rules were not written in advance; they were learned from a plausibility pre-check in which the language model read every dealt card and flagged the ones that could not be turned into a physically sensible scene. Each round of flags became a rule in the dealer, and the flag rate converged from 37 per 2000 cards to 4.

Writing and checking are separate calls. A writer model expands 40 cards per call into a caption plus a short scene summary and action summary. An independent validator re-reads every caption against the rules (exact adult count with a number word, full bodies with feet on the ground in the first second, flat visible named ground, one continuous take, one of the eight camera moves, 60 to 125 words, no minors, brands, text or weapons) and fixes or rejects. Code then checks length, forbidden words and near-duplicates across the whole set, comparing scene summaries by word-set similarity and captions within an action family, and every rejected card is rewritten with the reason attached. The result is 2000 captions with 2000 distinct places and no duplicates. The vocabularies and the full rule list are in Appendix~\ref{app:rules}, and the prompts in Appendix~\ref{app:prompts}.

\paragraph{Videos.} Each caption is sent to Veo 3.1 Fast~\cite{veo3} with the caption id as the seed: 6 seconds, 720p, 24 fps, no audio, 144 frames. The clip is resampled to the model's format, 81 frames at 16 fps, by dropping every third source frame and keeping the first 81 of the remaining ones, which preserves the anchored first second at the start of the clip.

\begin{figure*}[t]
  \centering
  \includegraphics[width=0.9\textwidth]{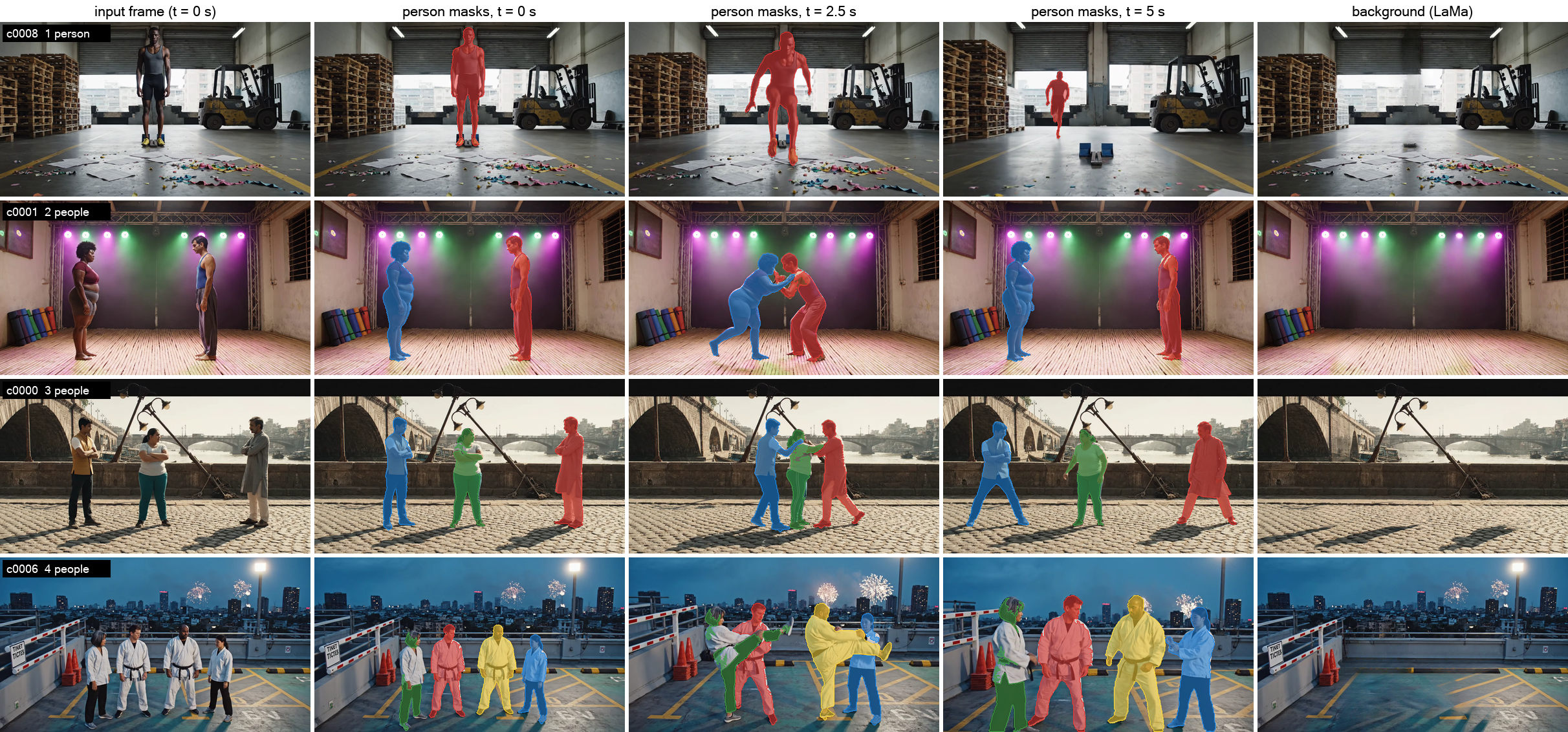}
  \caption{\textbf{Person tracking and background removal.} Rows: four clips with one to four people, indoor and outdoor, static and moving cameras. Columns: the input frame at $t=0$, the SAM 3.1 person masks at 0, 2.5 and 5\,s (one color per track), and the LaMa background obtained by inpainting the union of the masks in frame 0; this background is the reference image the model is conditioned on.}
  \label{fig:tracking}
\end{figure*}

\paragraph{Lifting clips to geometry.} Each clip is then reduced to the same objects the user will later draw (Figure~\ref{fig:engine}). People are tracked with SAM 3.1~\cite{sam3} video segmentation prompted with the word ``person'', which yields a mask, a box and a foot point per person per frame (Figure~\ref{fig:tracking}). The background image is frame 0 with the union of the person masks, dilated to cover contact shadows, inpainted by LaMa~\cite{lama}; this is the reference image the model is conditioned on, so the model learns that the reference shows the place and not the people. The ground mask is found by an agent loop on frame 0: a language model with vision sees the frame and the caption's ground words, proposes one short noun phrase for the walkable ground, SAM 3 segments that phrase, the model judges the overlay as accept, too little, too much or wrong surface, and revises the phrase, for at most three rounds. Cameras, a point map and surface normals come from HunyuanWorld-Mirror~\cite{worldmirror} run on 10 keyframes spread over the clip; poses for the other frames are interpolated linearly in translation and spherically in rotation. The ground plane is fitted by RANSAC to the reconstructed points that fall inside the ground mask and outside all person masks, with the reconstructed normals used twice: one-point hypotheses are drawn from a point and its normal, and inliers must agree with the plane normal within 30 degrees, which stops the fit from settling on a wall or on the people. The plane is cross-checked against an independent estimate: the plane's vanishing line, $l = K^{-\top} R\, n$ in the first camera, is compared with the horizon predicted from the single image by GeoCalib~\cite{geocalib}, and a disagreement above 60 pixels raises a flag. Each person's foot ray is intersected with the plane in every frame to give the cylinder's position; the height is the median over the first second of the foot-to-head extent projected onto the plane, and people who become fully visible later use their first visible second instead. The control video is then rendered from these cameras, this plane and these cylinders with the code of Sec.~\ref{sec:control}, so training and user-drawn controls share one renderer.

\paragraph{Quality control.} Every clip receives flags. Hard flags drop the clip: the ground loop failed three times (\texttt{ground\_fail}), the plane failed its sanity checks on inlier ratio, camera height or normal direction (\texttt{plane\_fail}), or no person had a usable anchor (\texttt{no\_anchor}). Soft flags are kept in the manifest for later decisions: the number of cylinders differs from the caption's count (\texttt{count\_mismatch}), a track jumps by more than one height in one frame (\texttt{track\_jump}), the anchor came from a later second (\texttt{anchor\_late}), the horizon disagrees with GeoCalib (\texttt{horizon\_disagree}), feet leave the frame later in the clip (\texttt{feet\_cut\_later}), the camera sits lower than 0.4 person heights (\texttt{low\_camera}), or at least five more people were found than the caption asked for (\texttt{bystanders}). On the full run of 2000 clips, 25 clips carry a hard flag (26 flags: 14 ground, 7 plane, 5 anchor; one clip carries two). Count mismatches are frequent (658 clips) but, on inspection, almost always the video model's doing rather than ours: in 77 clips it drew fewer people than the caption asked for, in 581 it drew more, mostly spectators who walked onto the playing surface; in both cases the cylinders match what is visible, so the flag is soft. The remaining 1975 clips are split into 1935 training tuples and 40 hold-out clips; the per-flag counts are in Appendix~\ref{app:qc}.

\begin{figure*}[t]
  \centering
  \includegraphics[width=0.9\textwidth]{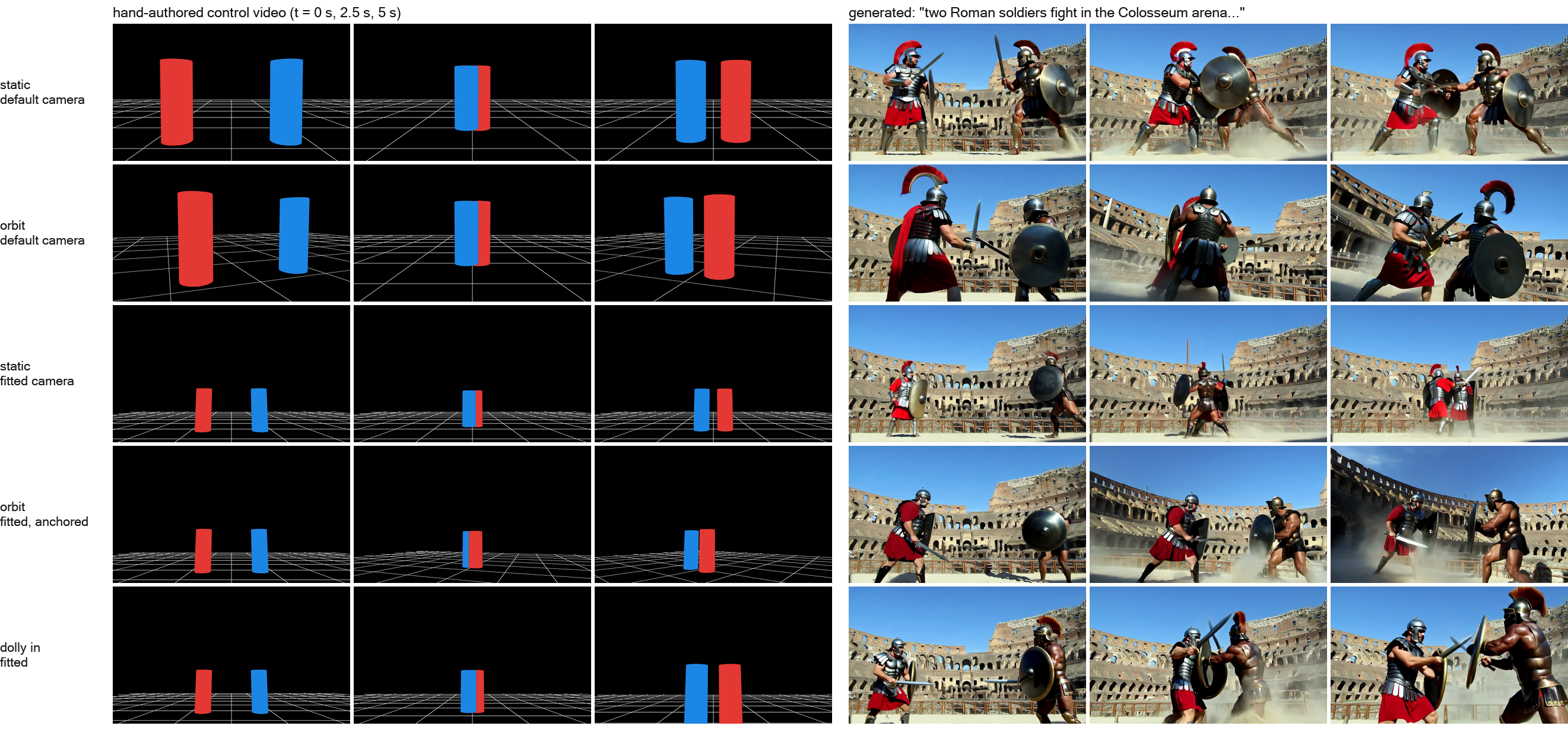}
  \caption{\textbf{Authoring from scratch, with and without fitting the ground of the photograph.} Two cylinders were placed in the editor, given curved paths that circle and swap sides, and exported under several camera presets (left); the background reference is a photograph of the Colosseum arena and the prompt reads ``two Roman soldiers fight with sword, spear and shields in the Colosseum arena''. With the default authoring camera (rows 1--2) the photograph's horizon lies 180 pixels below the grid's, and the fighters come out far larger than the cylinders prescribe. After fitting the camera to the photograph with GeoCalib (Sec.~\ref{sec:authoring}; rows 3--5) the fighters stand at the cylinders' distance with their feet on the sand, the dolly-in is followed, and the anchored orbit keeps the photograph's horizon in its first frame, although the model still enlarges the fighters towards the end of the orbit.}
  \label{fig:authoringfit}
\end{figure*}

\subsection{Training}
\label{sec:training}

The base model is Wan2.2-Fun-A14B-Control~\cite{videoxfun2025,wan2025}, a two-expert diffusion transformer (a high-noise expert and a low-noise expert) with a control branch that takes a control video and a reference path that takes one image. We use its \texttt{control\_ref} mode: the inputs are our control video, the LaMa background as the reference image and the caption; the target is the 81-frame clip.

One change to the upstream training code matters. Upstream takes the reference image from a frame of the training clip itself, which shows the people and therefore leaks the layout the control video is supposed to supply. We read the reference from the external background image instead and pass it through the same path the inference code uses when a reference image is given without a start frame (the first-frame channel zeroed, the reference carried by the full-reference self-attention path and by the image encoder), so training and inference agree.

We train LoRA adapters of rank 64 and alpha 32 on the query, key, value and both feed-forward projections of every block, one expert at a time, with the upstream loss. Clips are trained at the 480p bucket (token length 640) at their native 16:9 aspect, 81 frames, batch size one per GPU, AdamW with learning rate $10^{-4}$, weight decay $3 \times 10^{-2}$ and gradient clipping at 0.05, bf16 with gradient checkpointing, DeepSpeed ZeRO-2 on 8 H100 80GB GPUs. Each expert sees the 1935 tuples once: 241 steps at 27 seconds per step, about 1 hour 50 minutes per expert and 70 GB of memory per GPU. The upstream augmentations, 10 percent caption dropout, 10 percent zeroing of the reference and random inpainting masks on the target, are kept. A single-GPU smoke run showed why the one epoch is needed: after 100 samples the model still copied the cylinders literally, as a red block and a dark block at the right positions on the right background; after the full epoch they are people. At inference we use 50 steps, classifier-free guidance 6, LoRA weight 0.55 on both experts as upstream recommends, and render 81 frames at $480 \times 832$; we did not try higher resolutions.

\subsection{Authoring and inference}
\label{sec:authoring}

The authoring tool is deliberately minimal: a single web page (Fig.~\ref{fig:editor} in the appendix). The user places one to six people on a top-down view of the grid, sets each person's height, and drags them to new positions at keyframes; between keyframes the path is a cubic B\'ezier curve through the keyframe points (Catmull--Rom tangents by default, editable handles), sampled at constant speed within each segment. A camera panel offers the six presets used in our experiments (static, dolly in, dolly out, orbit, pan, crane) with editable height, pitch, yaw and focal length, and a live perspective preview of the grid and cylinders; the defaults are the medians of the training cameras (height 0.69 subject heights, pitch $2^\circ$, focal length 966 px at 1280 wide). The page exports a JSON scene, and a small script renders it with the same code that rendered the training data, so authored control videos sit inside the training distribution by construction. The user then picks a background image (a frame of a training clip with the people removed, a photograph, or a generated still) and writes a prompt; the trained model produces the video. Nothing in this loop needs a source video, a pose or a 3D model of a person.

\paragraph{Fitting the ground of a photograph.} In training, the control video and the background image come from the same clip, so the grid's horizon and the background's ground agree. A photograph brought in from outside has its own camera, and if the authored camera ignores it the model receives two contradictory grounds: in our first Colosseum test (Fig.~\ref{fig:authoringfit}) the photo's horizon lay 180 pixels below the grid's, and the fighters came out larger and closer than the cylinders prescribed. We therefore fit the ground of the photograph before authoring: GeoCalib~\cite{geocalib} estimates the focal length and the gravity direction from the single image, which fixes the horizon; the editor takes the focal length and the tilt below the horizon from this fit, draws the photograph under its preview together with both horizon lines, and leaves the camera height, in subject heights, to the user, who lowers or raises it until the cylinders' feet sit where people would stand. Camera presets that re-aim the camera at the people would move the horizon away from the photograph's in the first frame, so under a fitted camera the presets are replaced by rigid moves that start exactly at the fitted pose: an orbit about the vertical through the group, a pan, and a dolly. Figure~\ref{fig:authoringfit} compares the same authored duel with and without this fit: without it the fighters come out far larger than the cylinders prescribe; with it they stand at the cylinders' distance under a static camera and a dolly, while the anchored orbit keeps the photograph's horizon in its first frame but still lets the fighters grow towards the end.

\section{Experiments}
\label{sec:experiments}

We report dataset statistics, qualitative results on hold-out clips and on authored controls, failure cases and stress tests. We ran no human study, no ablation grid and no quantitative controllability measurement; the section carries its weight with breadth of results and with stress tests that point to concrete future directions, and the measurement protocol we would use is stated at the end.

\begin{table}[t]
  \centering
  \small
  \caption{\textbf{Dataset statistics.} Caption numbers are counted over the 2000 seed cards and captions; clip numbers over the full preprocessing run. A clip is dropped only when the ground mask, the plane fit or the anchor second fails; a mismatch between the caption's person count and the tracked count is kept as a soft flag.}
  \label{tab:dataset}
  \begin{tabular}{@{}lr@{}}
    \toprule
    Item & Count \\
    \midrule
    Captions = generated clips & 2000 \\
    Distinct (region, ground archetype) pairs & 2000 \\
    Regions / ground archetypes & 50 / 162 \\
    Action families (defined / used) & 118 / 116 \\
    Camera moves (each dealt 246--254 times) & 8 \\
    People per caption (1 to 6, Tab.~\ref{tab:axes}) & 1--6 \\
    Indoor / outdoor scenes & 613 / 1387 \\
    Words per caption, median (range) & 118 (94--125) \\
    Duplicate captions / duplicate scenes & 0 / 0 \\
    Clips passing ground, plane and anchor checks & 1975 \\
    Training / hold-out clips & 1935 / 40 \\
    Training clips with a moving camera & 1019 (53\%) \\
    \bottomrule
  \end{tabular}
\end{table}

\paragraph{Dataset statistics.} Table~\ref{tab:dataset} summarizes the training set. One measurement is worth stating here because it shapes what the model can learn: the caption asked for a moving camera in seven of eight cards, but the video model often ignored the instruction. Measured on the reconstructed cameras, 47 percent of the training clips have a static camera; crane (91 percent of its clips move), dolly-out (71 percent) and dolly-in (63 percent) were mostly honored, pan (56 percent) and orbit (46 percent) about half the time, and the sideways track (31 percent) and the handheld circle (40 percent) rarely. We did not re-weight moving clips.

\paragraph{Qualitative results.} Figure~\ref{fig:teaser} shows the central result on a hold-out clip: two cylinders on a stadium plaza, rendered once, produce two dancers who stand where the cylinders stand and keep their left-to-right order and height; changing only the prompt replaces the two dancers with an elderly man in tweed and a young woman in a yellow raincoat in the same places; changing only the reference image moves the same two dancers onto a frozen lagoon, where the model adds skates without being asked. On a second hold-out clip with three skaters the count, order and positions are again correct, and the same two edits, three different people by text and a stadium plaza by reference image, behave the same way. Figure~\ref{fig:camera} shows the same two cylinders under six authored camera paths: dolly-in is followed closely (the subjects grow into a close-up as the cylinders fill the frame), orbit shows the plaza from a rotated viewpoint by the end of the clip, crane produces an elevated view with the crowd rail below, and pan sweeps the background; dolly-out is followed only weakly, the subjects shrink less than the cylinders do. Motion inside a clip follows the caption (dance, skate) rather than copying the source clip's choreography, which is what a layout signal without pose should do. Figure~\ref{fig:holdout} has more results across prompts, geometries, counts and grounds.

\paragraph{Failure cases.} The failures we have seen so far are of six kinds (Figure~\ref{fig:failures}). Under a pan, one clip shows a third person in its first frames although two cylinders were drawn; count errors appear under moving cameras, not under static ones. Dolly-out is under-followed, consistent with dolly-out being one of the rarer honored moves in the training captions. Words in the prompt that the geometry never mentions still add objects: the camera sentence ``static tripod camera'' put a tripod on the ground during tai chi, and ``duck blind'' put a duck on the ice. Fast motion occasionally produces limb and pose glitches, and a snow spray in one skating clip turns into an amorphous blob for a few frames. A photograph whose viewpoint differs from the authored camera makes the people come out at the wrong scale, which is the failure the ground fit of Sec.~\ref{sec:authoring} removes. The generated videos inherit the letterboxing of the reference image when the source clip had black bars: about 6\% of the source clips are letterboxed (64-pixel bars at 720p), the background image keeps the bars and the model reproduces them; for the figures and videos in this paper we replace such a reference by a center crop and re-render the control video with the matching zoom, which leaves the geometry unchanged.

\paragraph{Stress tests and future directions.} Figure~\ref{fig:stress} in the appendix pushes one factor at a time outside the training range on hand-authored geometry. Count, occlusion order, crossings, an extreme height ratio and long or wide camera moves in the plane of the training cameras are followed; two things are not. First, viewpoints far from the reference image's (a steep top-down camera, a camera on the ground, a crane starting high above a photograph) are ignored: the reference image fixes the first-frame viewpoint and the control video can only move the camera away from it, which is why the ground of a photograph has to be fitted before authoring (Sec.~\ref{sec:authoring}). Second, when the text contradicts the geometry the model compromises rather than obeying either: fewer people than cylinders, or seated people under standing cylinders, at the cylinders' footprints. Each limitation of the representation is paired with the direction it points to. The cylinders carry no pose or heading, so the model chooses the action from the text; a heading mark or a stick figure inside the cylinder would give the user that control back. There is one ground plane, so stairs, terraces and slopes across the frame are out of scope; several planes or a height field would extend the signal. Cylinders stand on the ground, so jumps are approximated by a grounded cylinder; a per-frame height would model them. The training distribution is the text-to-video model's, with its own biases in people, places and camera habits (the static-camera share above is one); real footage lifted by the same engine would broaden it. Finally, the evaluation here is qualitative. The quantitative protocol we leave for future work measures four things on the hold-out clips and on authored controls: the foot-point reprojection error (SAM 3 on the generated video, feet lifted with the authored camera and plane, compared with the cylinder positions in grid cells), subject-count accuracy, camera consistency (the generated video reconstructed with the same model as the training data and its camera path compared with the authored one after alignment), and a reference-free video-quality score to check that the LoRA does not degrade the base model; the natural baselines are the base control model fed a depth or pose rendering of the same cylinders, and text alone.

\section{Related Work}
\label{sec:related}

\paragraph{Video backbones and control adapters.} Open video diffusion transformers, CogVideoX~\cite{cogvideox2024}, HunyuanVideo~\cite{hunyuanvideo} and Wan~\cite{wan2025}, now approach closed models such as Sora~\cite{sora} and Veo~\cite{veo3} in quality. ControlNet~\cite{controlnet} established the pattern of a trainable branch that injects a dense spatial condition into a frozen diffusion model; VideoComposer~\cite{videocomposer} and SparseCtrl~\cite{sparsectrl} brought it to video, and Fun-Control~\cite{videoxfun2025}, VACE~\cite{vace} and Ctrl-Adapter~\cite{ctrladapter} accept depth, pose, edge or tracked-video conditions for open backbones. Low-rank adaptation~\cite{lora} makes a new modality cheap to teach: our control video enters Fun-Control unchanged; a LoRA learns what cylinders on a grid mean.

\paragraph{Human motion control.} Pose-guided human video, Animate Anyone~\cite{animateanyone}, MagicAnimate~\cite{magicanimate}, Champ~\cite{champ} and Wan-Animate~\cite{wananimate}, animates a reference person from a skeleton or a parametric body, and Uni3C~\cite{uni3c} adds a camera path to SMPL-X bodies. All of them demand a full pose sequence, usually taken from a video the user already owns. A cylinder carries no pose: the model takes the motion from the text, and the user decides where the person stands.

\paragraph{Trajectory and motion control.} DragNUWA~\cite{dragnuwa}, MotionCtrl~\cite{motionctrl}, DragAnything~\cite{draganything}, Tora~\cite{tora} and Motion Prompting~\cite{motionprompting} move image regions or tracked points along 2D paths, Go-with-the-Flow~\cite{gowiththeflow} warps the noise along an optical flow, and Direct-a-Video~\cite{directavideo}, MotionCanvas~\cite{motioncanvas} and ObjCtrl-2.5D~\cite{objctrl25d} add depth or a camera to the paths. The paths still describe pixels: a person walking towards the camera and a person growing in place produce the same 2D track. Our cylinders stand on a ground plane the model can see, so a path means the same place under any camera.

\paragraph{Camera control.} CameraCtrl~\cite{cameractrl}, MotionCtrl~\cite{motionctrl} and AC3D~\cite{ac3d} condition on camera parameters, CameraCtrl II~\cite{cameractrl2} extends the control to dynamic scenes, GEN3C~\cite{gen3c} renders a 3D cache from the target camera, and ReCamMaster~\cite{recammaster} and TrajectoryCrafter~\cite{trajectorycrafter} re-render a given video from a new path. They steer the viewpoint but not the people, and they learn from camera-annotated footage such as RealEstate10K~\cite{realestate10k} or from synthetic scenes. In our representation the camera and the people are drawn in the same scene, and the grid makes the camera path visible to the model as image motion of the ground.

\paragraph{Layout and 3D-aware control.} GLIGEN~\cite{gligen} and Boximator~\cite{boximator} place 2D boxes in an image or over time, which loses the ground and the depth order that our plane restores. LooseControl~\cite{loosecontrol} conditions an image model on boxes and a ground plane rendered as depth, the closest representation to ours; Diffusion as Shader~\cite{das} renders 3D tracking videos as the condition, Perception-as-Control~\cite{perceptionascontrol} renders 3D-aware object and camera motion, and 3DTrajMaster~\cite{trajmaster3d} injects 6-DoF entity poses learned from a game-engine corpus. CineMaster~\cite{cinemaster} and VerseCrafter~\cite{versecrafter} come closest in scope, 3D boxes or Gaussian trajectories plus a camera, trained on automatically annotated real video, and the contemporaneous LooseControlVideo~\cite{loosecontrolvideo} conditions a Wan backbone on oriented 3D boxes that encode size, orientation and occlusion order, trained on stock footage. We ask for less: a cylinder has a footprint, a height and a color, which a person can draw in a minute. We train on a corpus generated from captions rather than on real footage, on an open model, and release the model, the data and the editor.

\paragraph{Data engines and previsualization.} Training a conditional model on pairs produced by other models has precedent: InstructPix2Pix~\cite{instructpix2pix} generated its editing pairs with a language model and an image model. Synthetic renderers such as Kubric~\cite{kubric} and the game-engine corpus of 3DTrajMaster~\cite{trajmaster3d} give exact geometry but not photoreal people; VerseCrafter~\cite{versecrafter} annotates real video instead. We do neither: a text-to-video model generates the footage from captions that guarantee coverage, and the geometry is recovered from it. The authoring loop resembles previsualization tools that pair a rough 3D scene with a generative model, CinePreGen~\cite{cinepregen} and PrevizWhiz~\cite{previzwhiz}, but it needs no game engine and no source video.

\paragraph{Scene understanding tools.} The engine depends on video segmentation with concepts (SAM 2 and SAM 3~\cite{sam2,sam3}), inpainting (LaMa~\cite{lama}), feed-forward multi-view reconstruction with normals (HunyuanWorld-Mirror~\cite{worldmirror}, in the family of VGGT~\cite{vggt} and Depth Anything 3~\cite{da3}) and single-image calibration (GeoCalib~\cite{geocalib}). None of them is modified; the contribution is the way they are combined and checked against each other.

\section{Conclusion}
\label{sec:conclusion}

A ground plane and one cylinder per person are enough to tell a video model where people stand, how they move and where the camera goes, while text and a reference image decide the look. The training pairs for such a signal can be built without real footage or manual labels, by generating videos from captions that guarantee coverage and lifting them back to their geometry. What this enables is simple: a person can author a multi-person shot with a camera move in minutes, from a sketch rather than a capture. Richer primitives (a heading, a stick figure, vehicle boxes), several planes, and real footage run through the same engine are the next steps~on~this~path.

{
    \small
    \bibliographystyle{ieeenat_fullname}
    \bibliography{main}
}

% Appendices A-E.
\clearpage
\section*{Appendix}

\appendix

\section{Caption rules and axis vocabularies}
\label{app:rules}

The hard rules the validator enforces, each with its downstream reason: (R1) main subjects counted exactly with a number word, adults only, no animals, because the cylinder count must equal the subject count; (R2) full bodies in frame with feet on the ground for at least the first second, because the cylinder needs an anchored foot point; (R3) flat, visible, named ground, never stairs or a mirror floor, because one plane is fitted; (R4) one continuous take, because one camera path is reconstructed; (R5) one of the eight camera moves; (R6) a stated motion range, with nobody leaving the frame; (R7) a still or a moving environment, stated; (R8) 60 to 125 words, English, photoreal, present tense; (R9) no minors, celebrities, brands, logos, readable text, weapons, nudity or gore; (R10) no template reuse across captions. The region, ground-archetype and action-family vocabularies with their eligibility tags are released with the code.

\begin{table*}[t]
  \centering
  \small
  \caption{\textbf{Caption axes.} Values = size of the vocabulary; dealing = how the sampler assigns the axis; measured = the count in the final 2000 seed cards (plan in parentheses where it differs). Balanced decks make shares exact; the small deviations come from the plausibility repair loop re-dealing a few cards.}
  \label{tab:axes}
  \begin{tabular}{@{}llll@{}}
    \toprule
    Axis & Values & Dealing & Measured \\
    \midrule
    Region & 50 & unique (region, archetype) pair & 2000 pairs \\
    Ground archetype & 162 & unique pair; tags pick decks & 162 used \\
    Action family & 118 & eligibility, least-used first & 116 used \\
    Camera move & 8 & uniform deck & 246--254 each \\
    Weather (outdoor) & 12 & deck, hostile capped 25\% & 1387 cards \\
    Lighting (indoor) & 6 & deck & 613 cards \\
    Environment motion & 14 + still & still 40\%, rest by venue & 800 still \\
    Motion range & 3 & 30 / 40 / 30 by venue size & 604 / 803 / 593 \\
    People & 1--6 & 500 / 450 / 350 / 300 / 200 / 200 & exact \\
    Crowd flag & -- & 20\% of audience venues & 398 (400) \\
    Close-approach flag & -- & 15\% of mobile actions & 299 (300) \\
    Vibe & 6 & uniform deck & -- \\
    Hold-out & -- & ids ending in 49 / 99 & 40 \\
    \bottomrule
  \end{tabular}
\end{table*}

\section{Agent prompts}
\label{app:prompts}

The prompts below are quoted verbatim from the code (\texttt{captions/capgen/prompts.py} in the repository) and from the preprocessing specification (\texttt{handoff\_mengyi/03\_ground\_and\_cameras.md}); only long vocabulary lists are elided with [...]. Every caption-engine call ran Claude Opus 5 (\texttt{claude-opus-5}) through Claude Code in print mode with a JSON schema for the output; the ground-mask agent ran the same model through the Anthropic API with the frame or the overlay attached as an image.

\paragraph{Shared rules.} The writer and the validator both receive this block as the start of their system prompt.
\begin{lstlisting}[style=prompt]
You write prompts for a text-to-video model (Wan 2.2, 5-second 720p clips at 16 fps). The captions build a training set for a model that learns: coarse 3D geometry (a ground plane plus one cylinder per main subject) + a background image + text -> photoreal video. Every caption must satisfy ALL rules.

R1 Main subjects: state the EXACT number of adult people with a number word, matching person_count. Describe them vividly (build, hair, outfit colors); vary gender, adult age range and ethnicity across captions. No animals anywhere.
R2 Other people: if crowd is false, nobody else is in the scene (empty background). If crowd is true, an audience or bystanders may appear but only OFF the playing surface - in stands, behind railings, on bleachers, in the far background - never standing on the same ground as the main subjects.
R3 Anchor: for at least the first second all main subjects are fully in frame with feet on the ground. If close_approach is false they stay fully in frame throughout. If close_approach is true, after the anchor one subject may move toward the camera and end partly out of frame - write that explicitly.
R4 Ground: name the ground material and keep the flat ground clearly visible under the subjects. A small area is fine. Wet, icy, snowy, shallow-water and gently sloped surfaces are fine. Never stairs, never a mirror floor.
R5 One continuous take: include a phrase like "one continuous shot"; never montage, cut, transition or scene-change wording.
R6 Camera: use exactly the given camera move, phrased naturally.
R7 Action and motion range: action_family is a DIRECTION, not a script - write one concrete action in the same spirit: a sibling move, a variation, a drill or a combination with the same body mechanics (for "judo hip throws" that could be a repeated o-goshi drill, alternating uchi-mata attempts, or a throw-and-roll sequence), so that captions sharing a family still show different specific moves. Use props only if the family implies them. in_place means the subjects stay on one spot (footprint under about 1 m); short_range means they move within a few meters; long_range means they cover ground, e.g. cross the frame. Within that range the action must be high-energy and physical, adapted to the person count and the venue size.
R8 Environment: if env_motion is "still", the environment is calm and only the people move. Otherwise weave the given motion element in vividly. Use the given lighting; if it is physically odd for the venue, adapt it plausibly (seen through windows or doors, or an equivalent indoor effect).
R9 Scene: invent ONE concrete place that fits region + archetype - not "a rooftop" but e.g. "the cracked helipad of a half-finished Dubai tower, idle cranes silhouetted behind". Give 2-3 distinctive visual details. Put a summary of at most 20 words in the scene field.
R10 Style: 60-125 words, English, present tense, photoreal; let the vibe set tone and color words. Forbidden: minors, real names or celebrities, brands, logos, readable text, real weapons (sport foils and wooden practice sticks are fine), NSFW, gore.
R11 Every caption must read differently from every other one: different openings, structure and details; never restate the seed fields as a list.
\end{lstlisting}

\paragraph{Writer.} Appended to the shared rules; the user turn is 40 seed cards (fields: id, person\_count, region, archetype, ground, indoor, size, lighting, env\_motion, motion range, action family, camera, crowd, close\_approach, vibe).
\begin{lstlisting}[style=prompt]
You receive seed cards as JSON lines. A card is a recipe, not a template: invent the concrete place and the concrete action from it. Return exactly one entry per card with the id copied verbatim, plus a scene summary (<= 20 words, the specific invented place) and an action summary (<= 12 words, what the people concretely do).
\end{lstlisting}

\paragraph{Validator.} Appended to the shared rules; the user turn is the (card, output) pairs of one batch.
\begin{lstlisting}[style=prompt]
You are the validator. You receive (card, output) pairs. Check every rule for each pair, especially: exact person count and no extra people unless crowd is true; the anchor and the close_approach behaviour; a named, visible, flat ground; the motion range actually written; a still environment when env_motion is "still"; the exact camera move; one continuous take; 60-125 words; forbidden content.
Verdict "ok" when compliant. Verdict "fixed" when the problems are minor: return the corrected caption (and corrected scene/action if they changed). Verdict "reject" only when the caption would need a full rewrite: give the reason.
Also compare the outputs in this batch with each other: if two scenes describe near-identical places or two captions share structure and phrasing, fix the later one so it clearly differs.
\end{lstlisting}

\paragraph{Repair.} Appended to the writer prompt when a card is re-dealt after a rejection or a near-duplicate; each card then carries a \texttt{why\_failed} field.
\begin{lstlisting}[style=prompt]
These cards failed a previous attempt. Each card carries a "why_failed" note; some quote a twin caption they were too similar to. Fix the cause: a clearly different place, a clearly different concrete action, or the rule that was broken. Follow every rule strictly.
\end{lstlisting}

\paragraph{Plausibility pre-check.} Run once over all 2000 seed cards before any writing; the flagged cards are re-dealt.
\begin{lstlisting}[style=prompt]
You review seed cards for a video-caption dataset before any writing happens. Each card fixes: region, ground archetype, lighting, environment motion, motion range (in_place / short_range / long_range), action family, camera, person count, crowd flag, close_approach flag.
The writer is allowed to adapt oddities (weather seen through windows, an indoor equivalent of an outdoor effect, a solo version of a group action). Flag ONLY cards that cannot be made plausible even with adaptation - for example a long-range sprint inside a space a few meters across, an action that physically needs equipment or space the venue cannot have, a crowd flag in a place where no audience could stand, or an action family that makes no sense on the surface (e.g. basketball drives on ice).
Do not flag merely unusual or creative combinations; those are wanted. Return only the flagged ids with a short reason.
\end{lstlisting}

\paragraph{Archetype tagging.} Run once over the ground archetypes to produce the eligibility tags the dealer uses (indoor, size, audience, surface); the output was reviewed by hand.
\begin{lstlisting}[style=prompt]
You tag ground archetypes for a video dataset sampler. For each archetype name, return:
- indoor: true if the location is under a roof (a stadium with open sky is outdoor; a hangar is indoor).
- size: "small" if the flat usable floor is under about 10 m across (boxing ring, squash court, dojo, veranda), "medium" for 10-40 m (tennis court, dance studio, plaza corner), "large" for over 40 m (runway, salt flat, stadium pitch, parking lot).
- audience_natural: true if an audience would plausibly watch from stands, bleachers, railings or a perimeter (sports venues, stages, arenas, plazas, tracks); false for workplaces, nature, infrastructure.
- surface: "ice" for ice surfaces, "water" for shallow-water or wet-sand surfaces, "reflective" for polished or wet floors that mirror light (marble, wet asphalt, epoxy), otherwise "normal".
Return every name verbatim, exactly once.
\end{lstlisting}

\paragraph{Ground-mask agent (proposer and judge).} The ground loop was implemented by our collaborator from the specification below, which fixes the content of both calls but not their exact wording; the per-clip logs (\texttt{ground\_agent.json}) record every proposed phrase, SAM 3 score, verdict and reason. Quoted from the specification (typographic characters replaced by ASCII):
\begin{lstlisting}[style=prompt]
Loop, at most 3 rounds:
1. Ask Claude for ONE short noun phrase for the flat walkable ground in this frame, given the image and the manifest `ground` words (e.g. `concrete floor`, `wet cobblestones`, `red clay court`). Tell it: 1-3 words, a surface not an object, no people.
2. Run SAM 3.1 with that phrase; union all masks with score >= 0.4; subtract the union of person masks (dilated 10 px) from step 1; drop connected components smaller than 0.5% of the frame.
3. Render an overlay (mask tinted red at 40% over the frame) and ask Claude to judge: does the red cover the flat ground the people stand on, and nothing else (no walls, furniture, sky, people)? Answer JSON: {"verdict": "accept" | "too_little" | "too_much" | "wrong_surface", "reason": "...", "new_phrase": "..."}.
4. Accept -> save. Otherwise use `new_phrase` and repeat. After 3 rejections: fallback mask = all non-person pixels in the lower 45% of the frame, `fallback: true`, and the clip is flagged `ground_fail` (still processed).
\end{lstlisting}
A typical accepted round from the logs: phrase \texttt{frozen lake ice}, SAM score 0.52, mask fraction 0.23, judge reason ``covers the frozen lake ice surface under and around the three people without spilling onto the reeds, hut, or sky''.

\section{Preprocessing quality control}
\label{app:qc}

Per-flag counts on the full run of 2000 clips are in Table~\ref{tab:qc}: hard flags on 25 clips (26 flags: ground 14, plane 7, anchor 5; one clip carries two); soft flags \texttt{count\_mismatch} 658, \texttt{track\_jump} 508, \texttt{anchor\_late} 338, \texttt{horizon\_disagree} 316, \texttt{feet\_cut\_later} 162, \texttt{bystanders} 141, \texttt{low\_camera} 113. Of the 658 count mismatches, 77 clips show fewer people than the caption asked for, 374 one or two more, and 207 three or more.

\begin{table*}[t]
  \centering
  \small
  \setlength{\tabcolsep}{6pt}
  \caption{\textbf{Preprocessing QC over the 2000 clips.} Hard flags drop a clip; soft flags are recorded in the training metadata. Horizon disagreement compares the horizon implied by the fitted plane with the one estimated from frame 0 by GeoCalib. Camera height is the fitted camera height divided by the median subject height.}
  \label{tab:qc}
  \begin{tabular}{@{}lrl@{}}
    \toprule
    Check & Clips & Note \\
    \midrule
    \multicolumn{3}{@{}l}{\emph{Hard flags (clip dropped)}} \\
    \texttt{ground\_fail} & 14 & no accepted ground mask \\
    \texttt{plane\_fail} & 7 & RANSAC plane rejected \\
    \texttt{no\_anchor} & 5 & no subject with feet on the plane \\
    Dropped (any of the three) & 25 & 1975 clips kept \\
    \midrule
    \multicolumn{3}{@{}l}{\emph{Soft flags (clip kept)}} \\
    \texttt{count\_mismatch} & 658 & tracked people $\neq$ caption count \\
    \texttt{track\_jump} & 508 & a track jumps between frames \\
    \texttt{anchor\_late} & 338 & height taken outside the first second \\
    \texttt{horizon\_disagree} & 316 & plane vs. GeoCalib horizon $>$ 60 px \\
    \texttt{feet\_cut\_later} & 162 & feet leave the frame after the anchor \\
    \texttt{bystanders} & 141 & $\geq 5$ more people than the caption \\
    \texttt{low\_camera} & 113 & camera height $<$ 0.4 subject heights \\
    \midrule
    \multicolumn{3}{@{}l}{\emph{Plane fit (medians)}} \\
    Inlier ratio & 0.98 & RANSAC with normal agreement \\
    Normal vs. camera up & $3.2^\circ$ & \\
    Horizon disagreement & 21 px & 90th percentile 77 px (of 720) \\
    Camera height / subject height & 0.69 & \\
    \midrule
    Training / hold-out split & 1935 / 40 & from the 1975 kept clips \\
    \bottomrule
  \end{tabular}
\end{table*}

\section{The authoring editor}
\label{app:editor}

\begin{figure*}[t]
  \centering
  \includegraphics[width=0.85\textwidth]{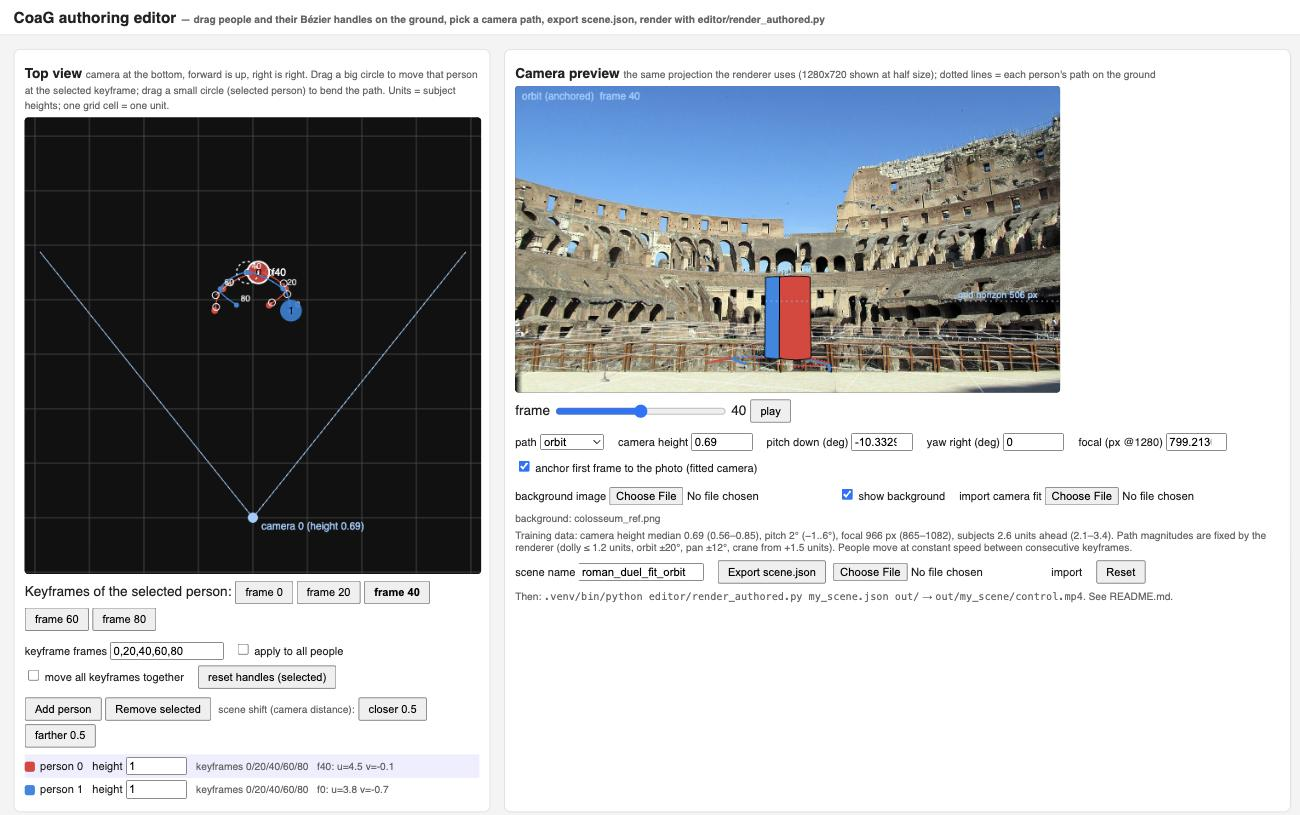}
  \caption{\textbf{The authoring editor.} Left: top view of the ground grid with the camera at the bottom; each person is a colored disc dragged to keyframes, joined by a B\'ezier path with editable handles (here two fighters that circle and swap sides). Right: the perspective preview drawn with the same projection as the renderer, here over the background photograph; the dashed line is the photograph's horizon from the ground fit, the dotted line the grid's horizon, and the camera panel holds the fitted focal length and tilt, the camera height in subject heights, the path preset and the anchoring switch. Export writes a JSON scene that a script renders into the 81-frame control video.}
  \label{fig:editor}
\end{figure*}

Figure~\ref{fig:editor} shows the editor described in Sec.~\ref{sec:authoring}: the top view where people are placed and their B\'ezier paths edited, the camera panel with the presets, the fitted-camera controls and the anchoring switch, and the perspective preview drawn with the renderer's projection, here over a background photograph with the photograph's horizon and the grid's horizon overlaid. The page exports a JSON scene; a script renders it into the control video with the same code that rendered the training data.

\section{Additional results and failure cases}
\label{app:results}

\begin{figure*}[t]
  \centering
  \includegraphics[width=0.9\textwidth]{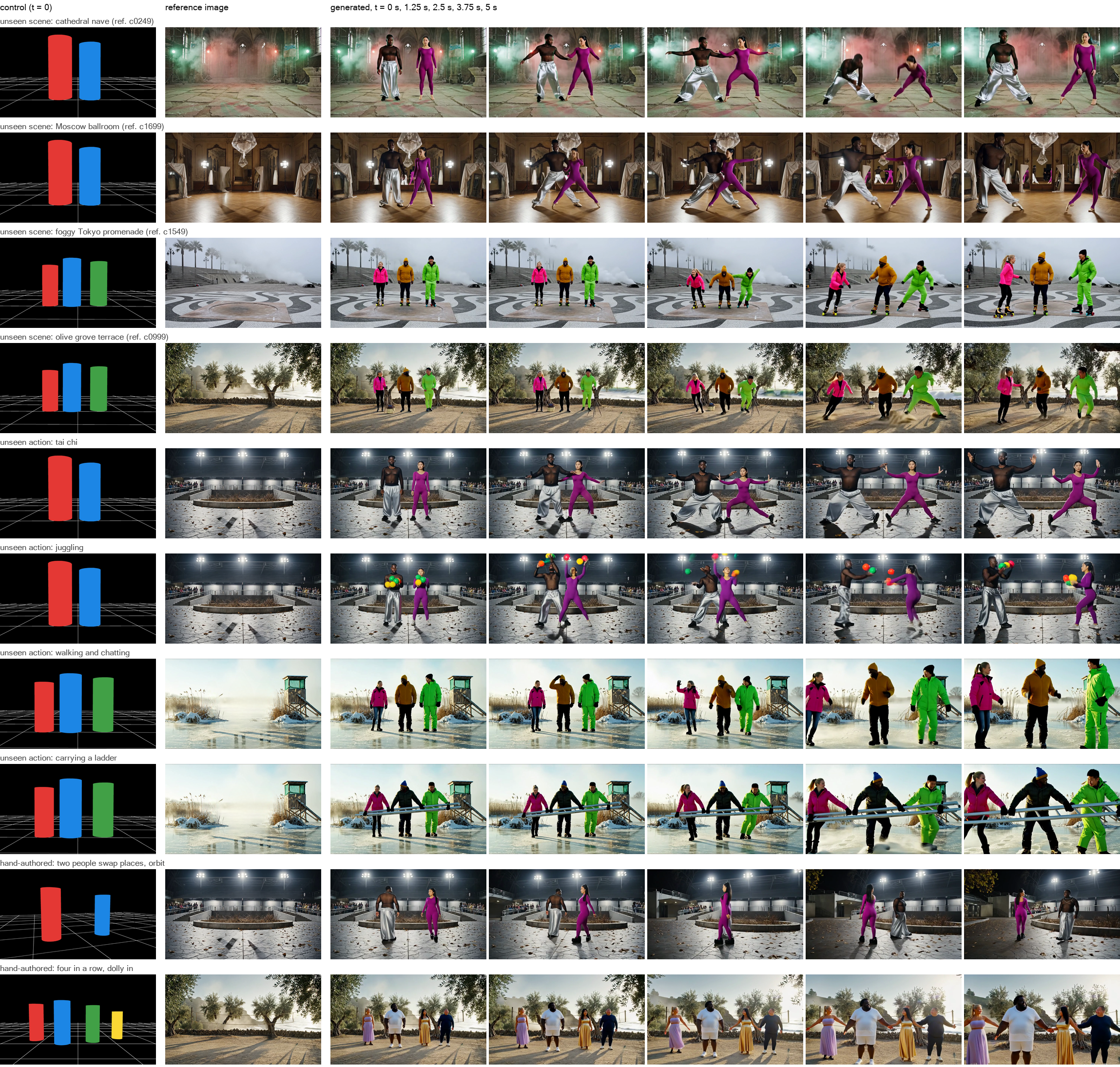}
  \caption{\textbf{Out-of-distribution tests, uniformly sampled frames.} Rows 1--4: the geometry of a hold-out clip with the background reference image of a different hold-out clip the model never saw. Rows 5--8: prompts with actions outside the 118 training action families. Rows 9--10: control videos authored by hand in the editor (Sec.~\ref{sec:authoring}), geometry that never occurred in training. Each row shows the first control frame, the reference image, and five frames of the generated video at 0, 1.25, 2.5, 3.75 and 5\,s. The full videos are on the project page: \url{https://zshyang.github.io/CoaG/}.}
  \label{fig:ood}
\end{figure*}

\begin{figure*}[t]
  \centering
  \includegraphics[width=0.9\textwidth]{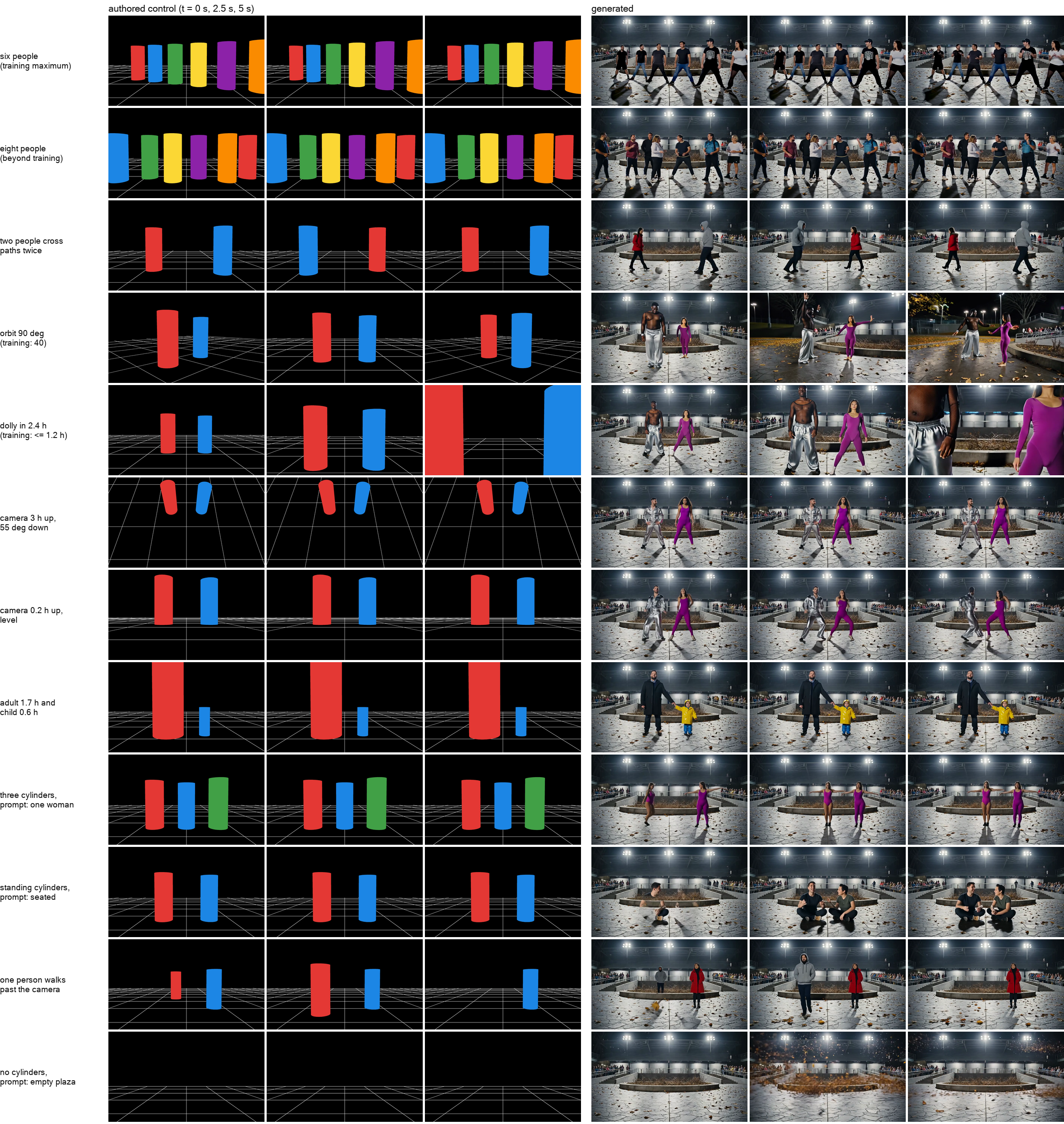}
  \caption{\textbf{Stress tests on hand-authored geometry} (same plaza background and static camera unless stated; rows in the order of the panel). Six people in a diagonal line come out as six; eight people, beyond the training range, come out as seven to eight with the palette cycling. Two people crossing paths twice swap sides at both crossings. A 90-degree orbit is followed by the background while the people hold their spots, and a dolly of 2.4 subject heights is followed all the way to a close-up; the steep top-down camera and the ground-level camera are not followed: the video stays at the eye-level viewpoint of the reference image. An adult of 1.7 units next to a child of 0.6 units keeps the height ratio. When the text contradicts the geometry the model compromises: three cylinders with the prompt ``one woman alone'' give two women, and standing cylinders with the prompt ``two people sit on the ground'' give two seated people at the cylinders' footprints, so the text decides the pose while the geometry decides the place. A person walking past the camera leaves the frame with the cylinder, and an empty control with the prompt ``empty plaza'' gives no people, though the ``leaves blowing'' clause fills the air with leaves.}
  \label{fig:stress}
\end{figure*}

Figure~\ref{fig:ood} shows uniformly sampled frames from three out-of-distribution tests run on hold-out geometry: background reference images from other hold-out clips (scenes the model never saw), prompts with actions outside the training action families, and control videos authored by hand in the editor. Full-length videos for every result in this paper, including the comparison set of Section~\ref{sec:experiments}, are on the project page (\url{https://zshyang.github.io/CoaG/}); the frames printed here are sampled at a fixed stride of 20 frames (1.25\,s) and are not selected by hand. Figure~\ref{fig:stress} collects the stress tests of Section~\ref{sec:experiments} on hand-authored geometry, again uniformly sampled; the full videos are on the project page.

\begin{figure*}[t]
  \centering
  \includegraphics[width=0.72\textwidth]{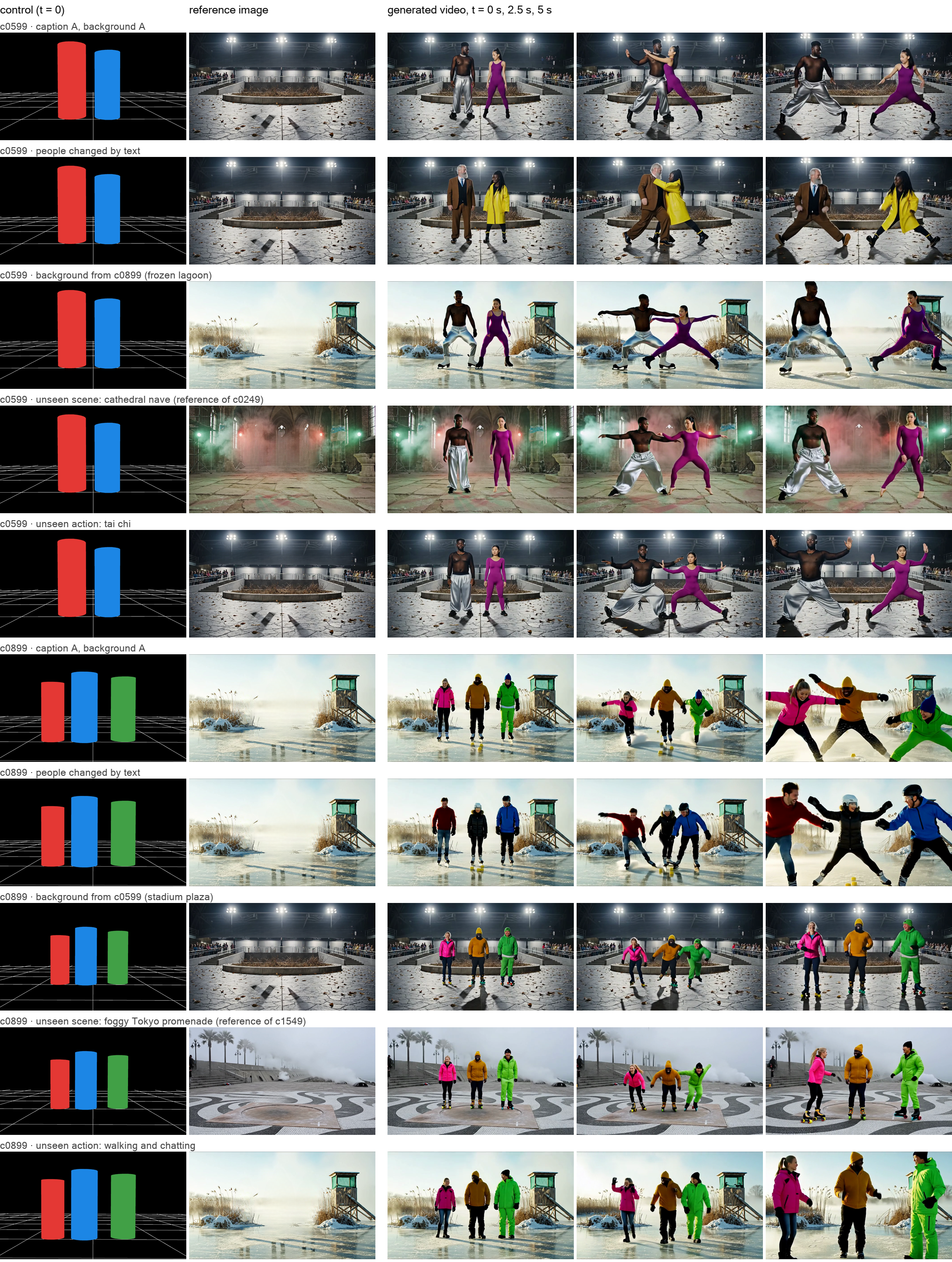}
  \caption{\textbf{Hold-out results.} Rows 1--5 share the geometry of hold-out clip c0599 (two people), rows 6--10 that of c0899 (three people); the first column is the first control frame, the second the background reference image, then three frames of the generated video. Within each block the rows change one thing at a time: nothing (the clip's own caption and background), the people in the text, the background reference image (taken from the other clip), the scene (a reference image from a third hold-out clip the model never saw), and the action (a prompt outside the training action families). Count, left-to-right order, footprint and height follow the cylinders throughout; appearance follows the text and the background follows the reference image. All clips are hold-out, one fixed seed, no per-case selection.}
  \label{fig:holdout}
\end{figure*}

\begin{figure*}[t]
  \centering
  \includegraphics[width=\textwidth]{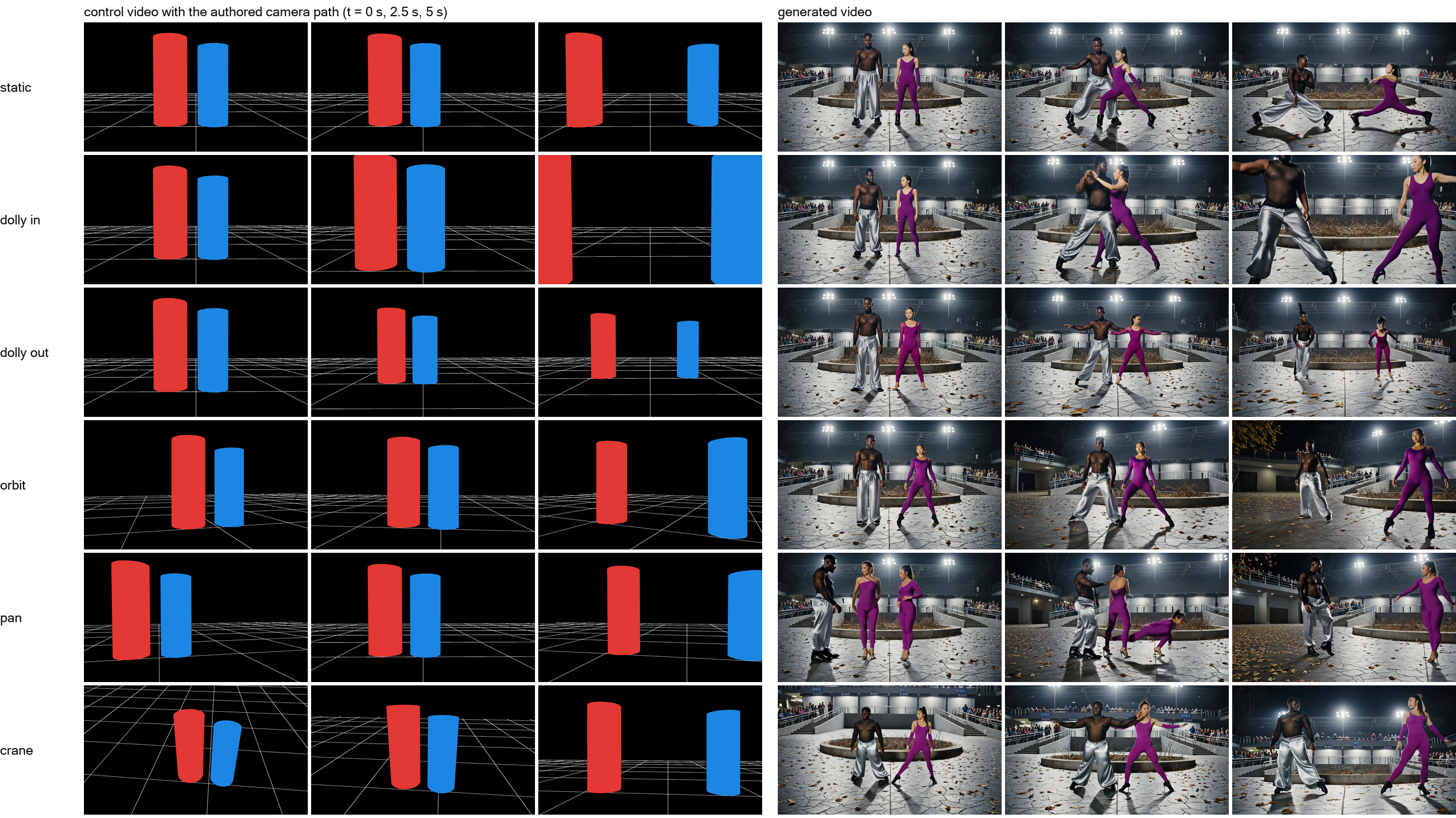}
  \caption{\textbf{Camera control.} The same two cylinders, caption and reference image rendered under six authored camera paths (left) and the resulting videos (right). Dolly in, orbit, pan and crane are followed; dolly out is followed only weakly (the subjects shrink less than the cylinders), and the pan clip shows a third person in its first frames.}
  \label{fig:camera}
\end{figure*}

\begin{figure}[t]
  \centering
  \includegraphics[width=\linewidth]{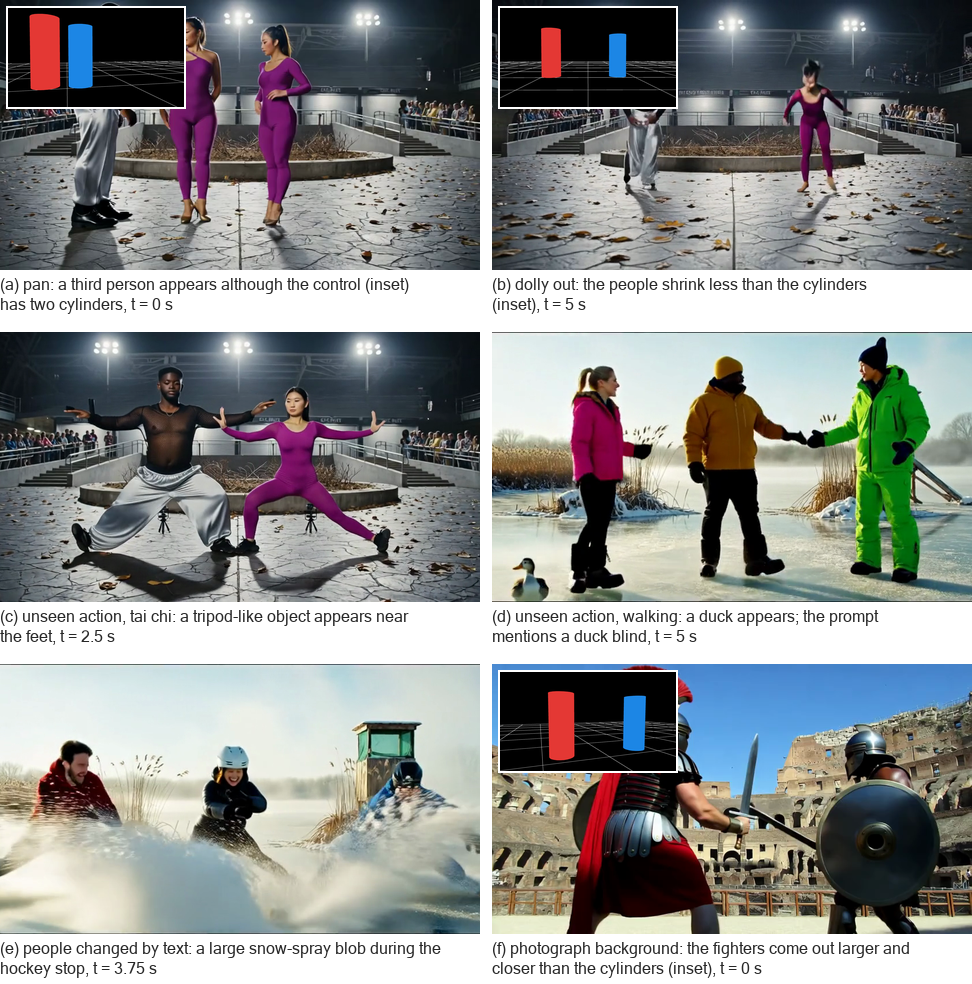}
  \caption{\textbf{Failure cases and what they point to.} (a) Under a pan a third person appears in the first frames although the control has two cylinders: count errors cluster under moving cameras, which are under-represented in training, so the first fix is more moving-camera data. (b) Under a dolly out the people shrink less than the cylinders, the weakest of the six camera paths; an explicit camera token or a camera-aware loss would give the model a second handle on the path. (c, d) Objects the geometry never mentioned appear when the text implies them (a tripod when the camera sentence says ``static tripod camera'', a duck after ``duck blind''): the text still adds content that the control does not constrain. (e) Fast motion still produces large artifacts (a snow-spray blob during a hockey stop). (f) With a photograph as background whose viewpoint is higher and wider than the training cameras, the fighters come out larger and closer than the cylinders prescribe; fitting the authored camera to the background image, as the data engine does for training clips, closes the gap (Fig.~\ref{fig:authoringfit}).}
  \label{fig:failures}
\end{figure}

\end{document}